\PassOptionsToPackage{table}{xcolor}

\documentclass[sigconf]{acmart}

\usepackage{multirow}
\usepackage{makecell}
\usepackage{colortbl}
\usepackage{pifont}          
\usepackage{enumitem}
\usepackage{subcaption}
\usepackage{placeins}        

\setcopyright{acmlicensed}
\copyrightyear{2026}
\acmYear{2026}
\acmDOI{XXXXXXX.XXXXXXX}
\acmConference[CIKM '26]{The 35th ACM International Conference on
  Information and Knowledge Management}{November 7-11, 2026}{Rome, Italy}
\acmISBN{978-1-4503-XXXX-X/26/11}

\title[TeachObs]{TeachObs: A Human-Validated Benchmark for Multimodal Teaching Observation and Model Evaluation}

\author{Yeil Jeong}
\email{yeilj@iu.edu}
\affiliation{%
  \institution{Indiana University Bloomington}
  \city{Bloomington}
  \country{USA}
}

\author{Youngjin Yoo}
\email{treplok@pcu.ac.kr}
\affiliation{%
  \institution{Pai Chai University}
  \country{South Korea}
}

\author{Jiyoung Bae}
\affiliation{%
  \institution{Korea University}
  \country{South Korea}
}

\author{Seobin Sohn}
\email{sbsohn5@snu.ac.kr}
\affiliation{%
  \institution{Seoul National University}
  \country{South Korea}
}

\author{Hyejin Han}
\email{hyereal@ewha.ac.kr}
\affiliation{%
  \institution{Ewha Womans University}
  \country{South Korea}
}

\author{Jinseo Lee}
\email{oesnij5313@gmail.com}
\affiliation{%
  \institution{Ewha Womans University}
  \country{South Korea}
}

\author{Howard Scott}
\email{howard.scott@wlv.ac.uk}
\affiliation{%
  \institution{University of Wolverhampton}
  \country{UK}
}

\author{Unggi Lee}
\email{codingchild@korea.ac.kr}
\affiliation{%
  \institution{Korea University Sejong Campus}
  \country{South Korea}
}

\renewcommand{\shortauthors}{Jeong et al.}

\makeatletter
\renewcommand{\@mkauthors}{%
  \gdef\@currentauthors{}%
  \gdef\@currentaffiliation{}%
  \gdef\@currentaffiliations{}%
  \global\setbox\mktitle@bx=\vbox{%
    \unvbox\mktitle@bx
    \par\medskip
    \hsize=\textwidth
    \centering
    {\LARGE Yeil Jeong\textsuperscript{1,\,*}, Youngjin Yoo\textsuperscript{2,\,*}, Jiyoung Bae\textsuperscript{3}, Seobin Sohn\textsuperscript{4,\,\textdaggerdbl}, Hyejin Han\textsuperscript{5,\,\textdaggerdbl},\\ Jinseo Lee\textsuperscript{5,\,\textdaggerdbl}, Howard Scott\textsuperscript{6}, Unggi Lee\textsuperscript{7,\,\textdagger}\par}
    \vspace{8pt}
    {\large\textsuperscript{1}Indiana University Bloomington, \textsuperscript{2}Pai Chai University, \textsuperscript{3}Korea University, \textsuperscript{4}Seoul National University,\\ \textsuperscript{5}Ewha Womans University, \textsuperscript{6}University of Wolverhampton, \textsuperscript{7}Korea University Sejong Campus\par}
    \vspace{8pt}
    {\large\textsuperscript{*}Co-first authors. \quad \textsuperscript{\textdaggerdbl}Equal contribution. \quad \textsuperscript{\textdagger}Corresponding author: codingchild@korea.ac.kr\par}
    \bigskip
  }%
}
\makeatother

\begin{document}

\begin{abstract}
Classroom videos contain observable teaching practices, but their pedagogical and visual signals are rarely organized in forms suitable for model evaluation.
We present \textit{TeachObs}, a human-validated benchmark for multimodal teaching observation in classroom videos.
\textit{TeachObs} includes 30 public lesson videos from eight countries divided into 5,158 fixed 15-second scenes.
Seven researchers annotated each scene with 39 binary observation codes, covering 20 visual codes, such as gesture, board work, pointing, and visual materials, and 19 nonvisual codes, such as instruction, monitoring, questioning, feedback, and reflection.
Gold segment labels are constructed using reliability- and prevalence-aware rules based on Krippendorff's alpha.
In addition to segment-level labels, three expert raters produced lesson-level ratings and qualitative evaluations of instructional design, instructional delivery, learner response, learning materials, and lesson closure across the 30 lessons, with rater coverage detailed in the body.
Using these two human reference layers, we evaluate five vision-capable frontier LLMs across three tracks - text-only segment coding, text + frame segment coding, and lesson-level coverage scored under an LLM-as-judge protocol - and find that no single model consistently outperforms others across all three tracks, that adding a mid-frame inflates both true and false attributions per scene, and that model evaluations over-rate procedurally clear lessons relative to expert raters.
\textit{TeachObs} therefore supports both fine-grained annotation benchmarking and whole-lesson evaluation, showing where AI systems can assist classroom video analysis and where expert judgment remains necessary across varied subjects, classroom formats, and annotation difficulty levels.
\end{abstract}



\maketitle

\section{Introduction}

\begin{figure*}[!t]
\centering
\includegraphics[width=\linewidth]{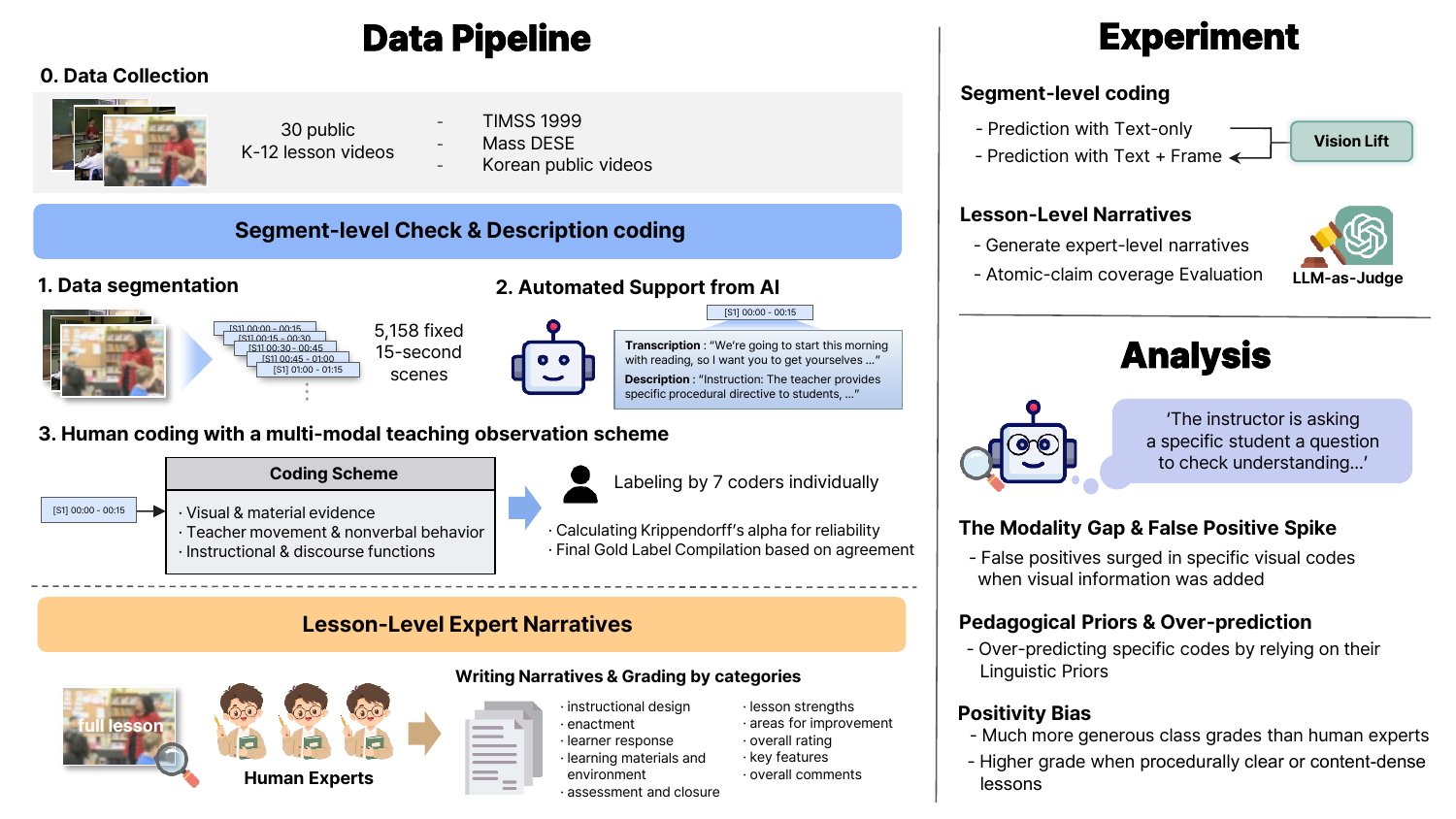}
\caption{Overview of \textit{TeachObs}. Two human-validated reference layers - segment-level multi-label codes and lesson-level expert narratives - share the same 30 lesson corpus and support three evaluation tracks that compare frontier LLMs along distinct axes (text-only segment coding, text + frame segment coding, and lesson-level coverage).}
\label{fig:overview}
\end{figure*}

What a teacher does in the classroom, and how well the resulting lesson works, are two questions human observers learn to answer through years of training.
The first question - \emph{which teaching actions were performed at a given moment} - is grounded in fine-grained, observable signals such as gestures, board work, and discourse moves.
The second question - \emph{how the lesson, as a whole, designed instruction and engaged learners} - calls for an integrated qualitative judgment over the entire session.
Building AI systems that approach either question requires reference data that explicitly separates the two, and which is annotated by trained human observers.

Existing classroom video resources fall short of this need on three fronts.
First, video datasets that include teaching footage are typically annotated for generic actions or for narrow domains~\citep{kay2017kinetics,mangalam2023egoschema}, leaving observable \emph{teaching} behaviors largely uncovered.
Second, work in educational NLP has concentrated on dialogue and tutoring text~\citep{macina2023mathdial}, with limited grounding in the visual channel of a real classroom.
Third, lesson-level qualitative observations from expert raters - the kind of writing that conventional teacher-observation rubrics produce - are rarely released alongside fine-grained segment annotations, so the two granularities cannot be jointly studied on the same lessons.

This paper introduces \textit{TeachObs}\footnote{\url{https://anonymous.4open.science/r/teacherOps-57D6/}}, a human-annotated benchmark over 30 public K-12 lesson videos.
\textit{TeachObs} releases two reference layers produced by trained human annotators (Figure~\ref{fig:overview}); a segment-level multi-label gold over 5{,}158 fixed 15-second scenes against a 39-item observation scheme (20 visual codes and 19 nonvisual codes), produced by seven researchers under a reliability- and prevalence-aware consensus procedure calibrated by Krippendorff's $\alpha$, and a set of lesson-level qualitative narratives across ten evaluation categories, produced by three expert raters whose per-lesson coverage is detailed in \S\ref{sec:narratives}.
The two layers cover the same lessons and are designed to be combinable, so that segment-scale behaviors and lesson-scale qualitative observations can be analyzed on the same recordings.

\paragraph{Contributions.}
\begin{itemize}[nosep,leftmargin=*]
  \item \textit{TeachObs}, a K-12 classroom benchmark with two human-validated reference layers on 30 lessons - 5{,}158 fifteen-second scenes against 39 observation codes, and lesson-level expert narratives across ten categories (\S\ref{sec:benchmark}).
  \item Three evaluation tracks (text-only and text + frame segment coding, plus lesson-level coverage) with headline numbers for five vision-capable frontier LLMs (\S\ref{sec:results}).
  \item Three triangulation findings on model failure modes - vision-driven over-attribution, pedagogical-discourse over-extension, and lesson-level over-rating - validated against independent expert re-judgment (\S\ref{sec:analysis}).
\end{itemize}

\section{Related Work}
\label{sec:related}

\newcommand{\cmark}{\textcolor{green!60!black}{\ding{51}}}
\newcommand{\xmark}{\textcolor{red!70!black}{\ding{55}}}
\newcommand{\bcmark}{\textcolor{green!60!black}{\ding{51}}}
\newcommand{\bxmark}{\textcolor{red!70!black}{\ding{55}}}

\begin{table*}[!t]
\centering
\footnotesize
\setlength{\tabcolsep}{2.5pt}
\caption{Position of \textit{TeachObs} relative to recent video VLM-evaluation benchmarks (Group 1), classroom data resources released for classifier baselines (Group 2), and the two classroom-domain VLM-evaluation benchmarks (Group 3). \emph{Modality} - V video, T transcript or dialogue, A audio. \emph{Subjects} and \emph{Countries} count distinct school subjects and source countries; "n/a" marks non-classroom resources. \emph{Grade} lists school-level coverage. \emph{Annotators} reports who produced the labels and how many people, where known. \emph{Teacher} marks whether the teacher is the labeled subject. \emph{Lesson narr.} marks whether free-text lesson evaluation written by human experts is part of the release (AI-generated or template-filled reports do not count).}
\label{tab:related-bench}
\begin{tabular}{@{}lllllllcc@{}}
\toprule
\textbf{Benchmark} & \textbf{Year} & \textbf{Modality} & \textbf{Size} & \textbf{Subjects} & \textbf{Countries} & \textbf{Annotators} & \textbf{Tch.} & \textbf{Narr.} \\
\midrule
\multicolumn{9}{@{}l}{\textit{Group 1.\ VLM-evaluation benchmarks (generic video)}} \\
Kinetics-700~\citep{kay2017kinetics}        & 2017 & V     & 650K clips                          & n/a       & n/a   & crowdworkers              & \xmark & \xmark \\
EgoSchema~\citep{mangalam2023egoschema}     & 2023 & V     & 5K MCQ over 250 h                   & n/a       & n/a   & human-curated             & \xmark & \xmark \\
MVBench~\citep{li2024mvbench}               & 2024 & V+T   & 4K QA, 20 tasks                     & n/a       & n/a   & auto + verified           & \xmark & \xmark \\
VideoMME~\citep{fu2024videomme}             & 2024 & V+T   & 2.7K videos, up to 1 h              & n/a       & n/a   & expert annotators         & \xmark & \xmark \\
MMBench-Video~\citep{fang2024mmbenchvideo}  & 2024 & V+T   & 600 videos, 2K+ QA                  & n/a       & n/a   & human-annotated           & \xmark & \xmark \\
E.T.\ Bench~\citep{liu2024etbench}          & 2024 & V+T   & 7K videos, 7.3K samples             & n/a       & n/a   & human-annotated           & \xmark & \xmark \\
\midrule
\multicolumn{9}{@{}l}{\textit{Group 2.\ Classroom data resources (not LLM-evaluation)}} \\
HowTo100M~\citep{miech2019howto100m}        & 2019 & V+T   & 100M clips                          & n/a       & n/a   & auto (narration)          & \xmark & \xmark \\
CIMA~\citep{stasaski2020cima}               & 2020 & T     & ~700 sessions                       & 1 (Botany) & 1    & crowdworkers              & \xmark & \xmark \\
EduNet~\citep{sharma2021edunet}             & 2021 & V     & 7.8K clips, 20 actions              & 1 (Gen.)  & 1    & research team             & \xmark & \xmark \\
NCTE Transcripts~\citep{demszky2023ncte}    & 2023 & T     & 1.7K lessons, 317 teachers          & 1 (Math)  & 1 (US) & trained raters (CLASS)  & \cmark & \xmark \\
SCB-Dataset~\citep{yang2023scb}             & 2023 & V     & 13K images, 19 classes              & 1 (Gen.)  & 1 (CN) & research team            & \xmark & \xmark \\
MathDial~\citep{macina2023mathdial}         & 2023 & T     & 2.9K sessions                       & 1 (Math)  & n/a   & teacher recruits          & \xmark & \xmark \\
MM-TBA~\citep{liu2025mmtba}                 & 2025 & V+T+A & 354 lessons, 3.2K clips             & 2 (Math, IT) & 1 (CN) & 4 undergrads (+LLM-gen.\ reports) & \cmark & \xmark \\
\midrule
\multicolumn{9}{@{}l}{\textit{Group 3.\ Classroom-domain VLM-evaluation benchmarks}} \\
SciIBI~\citep{shen2026sciibi}               & 2026 & V+T   & 113 clips                           & 1 (Sci)   & 1 (US) & education researchers      & \cmark & \xmark \\
\textbf{\textit{TeachObs} (ours)}           & \textbf{2026} & \textbf{V+T} & \textbf{30 lessons, 5{,}158 scenes, 1{,}030 claims} & \textbf{9 (Math, Sci, Eng, LA, SS, \ldots)} & \textbf{8} & \textbf{7 researchers} & \bcmark & \bcmark \\
\bottomrule
\end{tabular}
\end{table*}

\subsection{Classroom Video and Educational Annotation}

Most public classroom video corpora are organized around teacher observation frameworks such as Danielson's \emph{Framework for Teaching}~\citep{danielson_fft_review}, but their underlying segment-level annotations are rarely released.
Educational NLP work has produced dialogue and tutoring datasets~\citep{stasaski2020cima,macina2023mathdial} and the large-scale NCTE transcripts of elementary mathematics classrooms~\citep{demszky2023ncte}, yet the visual channel of a real lesson - boardwork, gestures, learning materials - is largely absent from these resources.
Recent classroom video work has begun to address the visual side; EduNet~\citep{sharma2021edunet} catalogues twenty teacher and student action classes in K-12 footage, SCB-Dataset~\citep{yang2023scb} detects student and teacher behaviors on still images, and MM-TBA~\citep{liu2025mmtba} combines short-clip teacher gesture annotation with four-dimensional lecture evaluation reports whose text content is generated by a fine-tuned LLM rather than authored by expert raters.
These are classroom data resources; they release labels for downstream classifiers but were not built around an LLM-evaluation protocol.
Adjacent K-12 educational benchmarks evaluate multimodal scientific or multi-discipline reasoning on exam-style questions rather than on classroom video~\citep{jiang2024visscience,zhou2025mdk12}.
The closest emerging neighbour we have located that is both a classroom video corpus \emph{and} an explicit LLM/VLM evaluation benchmark is SciIBI~\citep{shen2026sciibi} (arXiv preprint), which evaluates multimodal LLMs on Core Instructional Practice coding over 113 NGSS-aligned science clips.
\textit{TeachObs} sits in this same intersection and is the first to pair a multimodal scheme covering visual and discourse-level teaching with a lesson-level expert qualitative narrative layer on the same recordings.

\subsection{Multimodal and Long-Form Video Resources}

General multimodal video resources annotate action recognition, captioning, and question answering~\citep{kay2017kinetics,mangalam2023egoschema,fu2024videomme,zhou2024mlvu}, with recent benchmarks targeting comprehensive multi-task video understanding~\citep{li2024mvbench,liu2024tempcompass}, long-form holistic understanding~\citep{fang2024mmbenchvideo,wu2024longvideobench}, fine-grained temporal reasoning~\citep{cai2024temporalbench}, ultra-long egocentric video understanding~\citep{wang2025xlebench}, and open-ended event-level reasoning~\citep{liu2024etbench}.
These corpora do not target educational practice and treat the video as a generic event stream rather than as a structured teaching episode.
Work on instructional video has focused on procedural understanding in domains like cooking~\citep{miech2019howto100m}, which has different temporal and semantic structure from classroom teaching.
Table~\ref{tab:related-bench} positions \textit{TeachObs} against three groups of representative benchmarks - generic VLM-evaluation benchmarks, classroom data resources for downstream classifiers, and the only other classroom-domain VLM-evaluation benchmark.
The direct comparator is SciIBI, which evaluates LLMs on Core Instructional Practice coding over 113 clips in a single subject (science) and a single national context, annotated by the authoring research team.
\textit{TeachObs} sits in the same intersection but is broader along three independent axes - subjects (9 vs.\ 1), source countries (8 vs.\ 1), and school levels (kindergarten, elementary, middle, and high) - larger in scale by roughly a factor of forty-six on the segment side (5{,}158 scenes vs.\ 113 clips), and is the only resource in the comparison that combines a strict expert coding rubric with lesson-level expert narratives by trained raters on the same K-12 lessons.

\section{The \textit{TeachObs} Benchmark}
\label{sec:benchmark}

\textit{TeachObs} is a human-validated benchmark for multimodal teaching observation in classroom videos.
Unlike datasets that focus only on transcripts, student responses, or tutoring dialogue, \textit{TeachObs} represents classroom instruction at two complementary levels.
At the segment level, it provides fine-grained binary observation labels for fixed 15-second scenes.
At the lesson level, it provides expert ratings and qualitative evaluations of entire lessons.
Together, these two layers allow models to be evaluated not only on local evidence recognition - such as detecting gesture, board work, instruction, or feedback - but also on holistic teaching observation, such as judging instructional design, instructional delivery, learner response, materials, and closure.

We first identified publicly accessible classroom lesson videos from three source groups: the TIMSS 1999 Video Study archive, the Massachusetts Department of Elementary and Secondary Education (Mass DESE) calibration video resources, and Korean public classroom lesson videos.
The TIMSS 1999 Video Study is a large classroom video study of eighth-grade mathematics and science teaching across seven countries, involving videotaping and analysis of teaching practices in more than one thousand classrooms.
The Mass DESE source consists of public calibration videos developed for educator evaluation and classroom observation calibration.
The Korean videos were drawn from publicly accessible full-lesson classroom recordings.
Each video was divided into fixed 15-second scenes; transcripts and scene descriptions were produced through an automated video-analysis pipeline and used as structured supports for coding, not as final labels.\footnote{The released benchmark uses these automated artifacts only as coding aids; final segment-level labels were produced by independent human coding (\S\ref{sec:transcripts}-\S\ref{sec:gold}).}
Seven researchers with educational-technology backgrounds then independently coded all 5{,}158 scenes across all 30 videos, producing 210 coder-level annotation files ($7 \text{ coders} \times 30 \text{ videos}$).
These independent coder files were converted into 30 gold-label files through the reliability- and prevalence-aware aggregation procedure described in \S\ref{sec:gold}.
TeachObs therefore preserves a clear distinction among raw public videos, coder-level annotation files, and finalized gold benchmark files: the benchmark does not rely on automatically generated model labels as ground truth.

\subsection{Lesson Corpus and Scene Structure}
\label{sec:corpus}

\textit{TeachObs} contains 30 public classroom lesson videos selected from three source groups: 10 from TIMSS 1999, 10 from Mass DESE calibration videos, and 10 from Korean public classroom lesson videos.
The corpus spans eight countries (the United States, Korea, Japan, the Netherlands, the Czech Republic, Australia, Hong Kong, and Switzerland) and nine subject areas including mathematics, science, English, Korean language arts, ethics, social studies, integrated social studies, Korean geography, and an interdisciplinary English and social studies lesson.
School levels include kindergarten, elementary, middle, and high school.
Lesson lengths range from 20:18 to 55:47, with an average of approximately 43 minutes.

Each video was segmented into fixed 15-second scenes.
We chose fixed scenes rather than variable-length utterance units to support consistent multimodal model evaluation, temporal alignment, and reproducible scene-level coding.
The 30 videos yielded 5{,}158 scenes in total.
Each scene is represented by two paired rows in the source coder files: a \textit{Check} row and a \textit{Description} row.
The Check row contains binary True/False values for the 39 observation codes, while the Description row contains the corresponding textual description or rationale for the scene.
The full coder-level corpus therefore contains 5{,}158 Check rows and 5{,}158 Description rows, giving 10{,}316 paired data rows.
We use \textit{scene} to refer to the 15-second temporal unit and \textit{data row} to refer to either the Check or Description representation of that unit.
Our public gold release distributes the consensus Check labels per scene; the Description text is retained in the coder-level files used during reliability analysis and gold construction.
Table~\ref{tab:corpus} summarizes the corpus and scene-structure statistics.

\begin{table}[t]
\centering
\small
\setlength{\tabcolsep}{4pt}
\caption{TeachObs lesson corpus and scene-structure summary.}
\label{tab:corpus}
\begin{tabular}{@{}lr@{}}
\toprule
\textbf{Property} & \textbf{Value} \\
\midrule
Source groups & TIMSS 1999, Mass DESE, KR public \\
Videos per source group & 10 / 10 / 10 \\
Total videos & 30 \\
Countries & 8 (US, KR, JP, NL, CZ, AU, HK, CH) \\
Subject areas & 9 \\
School levels & 4 (kindergarten, elem., middle, high) \\
Duration range & 20:18 - 55:47 \\
Mean duration & $\sim$43 min \\
\midrule
Scene length & 15 s (fixed, non-overlapping) \\
Total scenes & 5{,}158 \\
Check rows & 5{,}158 \\
Description rows & 5{,}158 \\
Total paired rows & 10{,}316 \\
\midrule
Independent coders & 7 \\
Coder-level files & 210 ($7 \times 30$) \\
Observation codes & 39 (20 visual + 19 nonvisual) \\
\bottomrule
\end{tabular}
\end{table}

\subsection{Multimodal Teaching Observation Coding Scheme}
\label{sec:scheme}

TeachObs uses a 39-code binary multi-label coding scheme for multimodal teaching observation.
The scheme was not created solely through internal agreement among the researchers; instead, it was developed by adapting and operationalizing categories from prior work on teaching observation, instructional discourse, multimodal representation, and classroom interaction.
Specifically, the multimodal categories - including board work, instructional questioning, and organizational metadiscourse - build on \citet{tang2023multimodal}'s analysis of multimodal design in classroom explainer videos, while the dialogue-oriented categories - including review and prior-knowledge elicitation - build on \citet{hennessy2016seda}'s coding scheme for classroom dialogue analysis. Additional teacher-action categories, such as demonstration, movement, checking, cueing, responding, incorporating student ideas, and lecture, were adapted from prior classroom-observation work to the 15-second scene unit.
These references were used to transform theoretical and observational constructs into scene-level codes that can be applied consistently to classroom videos.

The resulting scheme is broader than a narrow teacher-action taxonomy.
We define it as a multimodal teaching observation scheme because many instructional signals in classroom videos are not purely verbal actions.
Some codes capture \emph{visual or material evidence}, such as Board work, Typeface, Mathematical-scientific, Drawing, Table, Graph, Map, Photograph, Video, Pointing, Underlining, Enclosing, Marking, and Linking.
Others capture \emph{teacher movement and nonverbal behavior}, including Moving, Stationary, and Gesture.
Still others capture \emph{instructional and discourse functions}, including Instruction (imperative classroom actions), Responding, Reinforcing, Checking, Cueing, Monitoring, Incorporating, Justifying, Review, Assessment, Reflection, Attention Orientation, Instructional Question (released as the \emph{Instructional} code), Rhetorical Question (\emph{Rhetorical}), Objective and Content Statements (\emph{Objective} and \emph{Content}), Prior Knowledge Elicitation, Organizational Metadiscourse, and Lecture.
The released coding scheme uses short code IDs (e.g., \emph{Instruction}, \emph{Instructional}) that the descriptive names above expand for readability; \emph{Instruction} marks imperative classroom actions while \emph{Instructional} marks instructional questions, and the two are distinct codes.

The 39 codes are divided into 20 visual codes and 19 nonvisual codes.
A single 15-second scene may carry multiple positive labels because classroom instruction frequently layers several forms of evidence; for example, a teacher may point to the board, provide an instruction, and monitor student responses within the same scene.
This multi-label structure allows TeachObs to represent the layered nature of classroom teaching more faithfully than a mutually exclusive label set.

The coding scheme was refined through collaborative development and training.
Coders jointly examined the literature-derived categories, discussed label definitions and boundary cases, and completed five rounds of training on classroom videos drawn from outside the 30-video target corpus.
Throughout coding, coders kept records of questions, ambiguities, and revision decisions; these records were used to maintain consistency across the full annotation process.
Full operational definitions, literature references, examples, and boundary rules for all 39 codes are released alongside the dataset.

\subsection{Automated Transcripts, Scene Descriptions, and Human Coding}
\label{sec:transcripts}

For each 15-second scene, transcripts and scene-level descriptions were generated before human coding.
The pipeline used Whisper large-v3 for non-English audio transcription, YouTube captions where available, and a frontier multimodal model for scene-level visual descriptions, all aligned to the 15-second segmentation (cf.\ \S\ref{sec:genai}).
These automatically generated artifacts were not treated as final annotations; instead, they served as structured inputs for human coders, who directly reviewed the videos while applying the observation codes.
The \textit{TeachObs} v0.1 release distributes the cleaned coder-revised transcripts (TSV) per scene, not the raw automatic output.

This distinction is important for the benchmark's integrity.
TeachObs does not use model-generated labels as ground truth.
The automated pipeline provided transcripts and visual descriptions to support efficient coding and standardized scene representation, but the final segment-level labels were produced through independent human coding and reliability-aware gold label construction.
Each of the seven coders independently coded all 30 videos, producing 210 coder-level TSV files.
Because every coder annotated every scene, the dataset supports a stronger analysis of inter-coder reliability than designs in which only a subset of scenes is double-coded.

Each coder-level file followed the same schema: seven metadata columns followed by 39 binary code columns.
The metadata columns identify the scene number, timestamp, filename, transcript, scene description, a blank separator column, and a row-type indicator.
The row-type indicator distinguishes Check rows from Description rows; Check rows store binary code decisions, while Description rows store scene descriptions or rationale-like text associated with the same 15-second unit.
This paired structure supports both conventional classification benchmarks and future analyses of model explanations or evidence grounding.

\subsection{Inter-Coder Reliability}
\label{sec:reliability}

We examined inter-coder reliability across all 30 videos and 39 codes.
Because all seven coders independently coded the full corpus, reliability could be examined at scale rather than on a small adjudication subset.
We used Krippendorff's $\alpha$ as the primary reliability coefficient.
Krippendorff's $\alpha$ is appropriate for estimating agreement among multiple coders and can be applied to nominal or binary decisions, making it suitable for evaluating how consistently the seven coders applied each observation code.
We computed reliability per code and used the results to guide the gold label construction procedure described in \S\ref{sec:gold}.

In addition to Krippendorff's $\alpha$, we report Gwet's AC$_1$ for rare codes, because $\alpha$- and $\kappa$-type indices can become unstable when positive labels are very sparse.
In teaching observation, low prevalence does not necessarily mean low importance: some instructional events such as assessment, reflection, or specific forms of feedback may occur rarely but remain educationally meaningful.
The reliability analysis therefore considered both agreement and prevalence rather than applying a single undifferentiated reliability threshold to all codes.

For the 23 codes with prevalence of at least 5\%, the median Krippendorff's $\alpha$ was 0.77, with values ranging from 0.37 to 0.90.
Across all codes, 48.7\% met the conventional $\alpha \geq 0.67$ criterion.
For rare codes with low prevalence, Gwet's AC$_1$ was additionally reported to address the prevalence problem, yielding a median AC$_1$ of 0.98.
These results indicate that many common observation codes reached acceptable inter-coder agreement, while some codes remained more difficult due to rarity, contextual dependence, or interpretive ambiguity.

We interpret this reliability pattern as part of the benchmark's contribution rather than merely a limitation.
Teaching observation is not a uniform recognition task: some codes correspond to directly visible or frequently occurring behaviors (e.g., \textit{Gesture}, \textit{Board work}, \textit{Instruction}), while others correspond to sparse but pedagogically meaningful events (e.g., \textit{Reflection}, \textit{Assessment}, certain forms of feedback).
A benchmark for teaching observation should preserve information about reliability and prevalence rather than collapse all labels into a single undifferentiated gold standard.
Per-code prevalence and reliability statistics are released with the annotation files.

\subsection{Reliability- and Prevalence-Aware Gold Label Construction}
\label{sec:gold}

We constructed segment-level gold labels using a reliability- and prevalence-aware aggregation procedure.
A single global majority-vote rule would be inappropriate for all 39 codes because the codes differ substantially in prevalence, observability, and interpretive difficulty.
Some codes are frequent and reliably identified by coders; others are rare but pedagogically meaningful; still others require interpretation of visual, material, or instructional context.
We therefore constructed gold labels using different aggregation criteria depending on prevalence, inter-coder reliability, and code type.

\paragraph{Low-prevalence codes (prev. $<$ 5\%).}
We used an \emph{inclusive consensus} criterion: if at least one coder marked the code as present, the gold label was set to positive.
This criterion was used because rare instructional events may be educationally meaningful even when only a small number of coders identify them; strict majority voting would erase many such low-frequency events.

\paragraph{High-prevalence, reliable codes (prev. $\geq$ 5\%, $\alpha \geq 0.67$).}
We used \emph{majority voting}.
In this condition the code occurs frequently enough and coders agree sufficiently, so a majority-based rule provides a conservative and stable gold label.

\paragraph{High-prevalence, lower-reliability action codes (prev. $\geq$ 5\%, $\alpha <0.67$, action).}
We used \emph{inclusive consensus}.
These codes often involve visually dependent or interpretively complex events, and lower agreement here may indicate that coders noticed different but plausible aspects of the same classroom event.
Inclusive consensus allows the gold labels to preserve these potentially valid observations rather than treating disagreement as absence.

\paragraph{High-prevalence, lower-reliability non-action codes (prev. $\geq$ 5\%, $\alpha < 0.67$, non-action).}
We used \emph{majority voting}.
These codes are more directly recoverable from transcript or explicit classroom discourse, so majority voting is used to avoid over-including ambiguous labels.

Applied to the full coder-level audit, this procedure covered 201{,}162 code cells in total and produced 42{,}756 final positive labels.
The per-condition breakdown is reported in Table~\ref{tab:gold-aggregation}.

\begin{table}[t]
\centering
\footnotesize
\setlength{\tabcolsep}{3pt}
\caption{Reliability- and prevalence-aware gold aggregation. Cells are taken over the gold release (5{,}158 Check rows $\times$ 39 codes = 201{,}162 cells; per-condition share 47.9 / 30.9 / 18.1 / 3.1\%). \emph{Incl.}\,= inclusive consensus; \emph{Maj.}\,= majority vote.}
\label{tab:gold-aggregation}
\begin{tabular}{@{}lcrr@{}}
\toprule
\textbf{Condition} & \textbf{Rule} & \textbf{Cells} & \textbf{Positives} \\
\midrule
$p<5\%$ & Incl. & 96{,}322 & 3{,}821 \\
$p\geq5\%$, $\alpha\geq.67$ & Maj. & 62{,}240 & 17{,}775 \\
$p\geq5\%$, $\alpha<.67$, act. & Incl. & 36{,}396 & 20{,}029 \\
$p\geq5\%$, $\alpha<.67$, non-act. & Maj. & 6{,}204 & 1{,}131 \\
\midrule
Total & & 201{,}162 & 42{,}756 \\
\bottomrule
\end{tabular}
\end{table}

This procedure is more transparent than a single global voting rule and preserves low-frequency codes that may be educationally consequential.
In classroom observation, rare events are not necessarily unimportant; they may represent critical moments of teacher noticing, student reasoning, instructional response, or classroom interaction.
We release the aggregation criteria as part of the benchmark metadata so future users can reproduce, audit, or modify the gold construction procedure.

\subsection{Lesson-Level Expert Narratives}
\label{sec:narratives}

In addition to the segment-level Check and Description coding, \textit{TeachObs} releases lesson-level expert narratives.
Three expert raters contributed lesson-level narratives across the 30 lessons; R1 and R2 rated all 30 lessons, while R3 rated 10 lessons.
They wrote qualitative observations across 10 categories: \textit{instructional design}, \textit{instructional delivery}, \textit{learner response}, \textit{learning materials and environment}, \textit{assessment and closure}, \textit{lesson strengths}, \textit{areas for improvement}, \textit{overall rating}, \textit{key features}, and \textit{overall comments}.
Eight of the ten categories also carry an ordinal rating (High / Mid / Low); the remaining two are narrative-only.

Narratives are stylistically diverse, ranging from paragraph-style prose to brief summaries to bullet lists.
\textit{TeachObs} releases the raw rater narratives in this form, together with the lesson and category metadata required to align them.

To enable lesson-level coverage scoring against these narratives, we decompose each rater narrative into a set of atomic factual or evaluative claims using a fixed-prompt decomposer (GPT-5.5; cf.\ \S\ref{sec:genai}).
Claims are deduplicated by category within each lesson; the test split contains 1{,}030 such atomic claims across the seven test lessons (4{,}045 across all 30 lessons).
The atomic-claim reference set, the decomposer prompt, and the per-lesson claim counts are released with the metadata.
A model narrative is scored against this reference by an LLM-as-judge protocol that decides, per claim, whether the model narrative semantically covers it (\S\ref{sec:results-setup}).

\section{Results}
\label{sec:results}

\begin{table}[!t]
\centering
\small
\setlength{\tabcolsep}{6pt}
\caption{Vision lift per model on the six-lesson intersection. Best value per column in \textbf{bold}, second best \underline{underlined}. Every model benefits from a single mid-frame image.}
\label{tab:vision-lift}
\begin{tabular}{@{}lcccc@{}}
\toprule
\textbf{Model} & 1-1 macro & 1-2 macro & $\Delta$ macro & $\Delta$ micro \\
\midrule
Claude Opus 4.7        & \textbf{0.277} & \underline{0.438} & +0.161 & +0.183 \\
GPT-5.5                & 0.254 & 0.435 & \underline{+0.181} & \underline{+0.212} \\
Gemini 3.1 Pro Preview & \underline{0.262} & \textbf{0.510} & \textbf{+0.247} & \textbf{+0.279} \\
Grok 4.3               & 0.251 & 0.413 & +0.162 & +0.196 \\
GLM-5V Turbo           & 0.236 & 0.378 & +0.142 & +0.170 \\
\bottomrule
\end{tabular}
\end{table}

We evaluate five vision-capable frontier LLMs on the three \textit{TeachObs} tracks introduced in \S\ref{sec:benchmark}.
Track 1-1 isolates the language channel by giving each model the scene transcript alone.
Track 1-2 keeps the same transcript and additionally attaches the scene mid-frame as a vision token (text + frame input), holding everything else fixed.
Track 2 asks each model to produce a lesson-level narrative across eight rated categories and scores the narrative against an atomic-claim reference set under an LLM-as-judge protocol.

\subsection{Setup}
\label{sec:results-setup}

\paragraph{Models.}
We report five vision-capable frontier LLMs that completed all three tracks.
The five are Claude Opus 4.7, GPT-5.5, Gemini 3.1 Pro Preview, Grok 4.3, and GLM-5V Turbo.
All models are accessed through the OpenRouter API with temperature 0 and low-effort reasoning where the model exposes such a control.

\paragraph{Evaluation split.}
The test split fixes seven lessons (S2, S4, S5, S19, S24, S28, S30) covering 1,312 scenes, about 25\% of the 5,158 scenes in \textit{TeachObs}.
S4 is omitted from Track 1-2 because its mid-frame attachments could not be produced uniformly for all five models in this run, so Track 1-2 and Track 1-1 are both reported on the six-lesson intersection (S2, S5, S19, S24, S28, S30) when used for modality comparison.
Track 2 is reported on all seven test lessons (1,030 atomic reference claims) because lesson-level inputs are not affected by per-scene frame extraction.

\paragraph{Metrics.}
Track 1 uses macro F1 (unweighted mean over the 39 codes) and micro F1 (pooled across scenes and codes).
Track 2 uses reference coverage; for each reference atomic claim, an LLM judge decides whether the model narrative semantically covers it, and we report the per-lesson macro mean ($\text{cov}_M$) and the micro mean pooled over all reference claims ($\text{cov}_\mu$).
The Track 2 judge in this version is \texttt{openai/gpt-5-mini} in single-judge mode; a family-disjoint three-judge majority is scheduled and is discussed in \S\ref{sec:limitations}.

\paragraph{Prompts and decoding.}
All models receive the same English prompt and are required to return strict JSON (codes for Track 1, eight-category rating, narrative, and reason for Track 2), with an additional free-text \texttt{reason} field per output.
Transcripts are kept in the lesson's source language, while the prompt and the requested output are English, so the benchmark also measures multilingual transfer.

\subsection{Headline Performance}
\label{sec:results-headline}

\begin{table*}[!t]
\centering
\small
\setlength{\tabcolsep}{6pt}
\caption{Headline performance of five vision-capable frontier LLMs on the three \textit{TeachObs} tracks. Track 1-1 and Track 1-2 are reported on the six-lesson intersection (S2, S5, S19, S24, S28, S30, $n{=}1{,}099$ scenes); Track 2 is reported on all seven test lessons (1,030 atomic reference claims). Best value per column in \textbf{bold}, second best \underline{underlined}.}
\label{tab:headline}
\begin{tabular}{@{}lcccccc@{}}
\toprule
& \multicolumn{2}{c}{\textbf{Track 1-1 (text)}} & \multicolumn{2}{c}{\textbf{Track 1-2 (text + frame)}} & \multicolumn{2}{c}{\textbf{Track 2 (coverage)}} \\
\cmidrule(lr){2-3}\cmidrule(lr){4-5}\cmidrule(lr){6-7}
\textbf{Model} & macro F1 & micro F1 & macro F1 & micro F1 & $\text{cov}_M$ & $\text{cov}_\mu$ \\
\midrule
Claude Opus 4.7        & \textbf{0.277} & \textbf{0.387} & \underline{0.438} & 0.570 & \underline{0.295} & \underline{0.293} \\
GPT-5.5                & 0.254 & 0.361 & 0.435 & \underline{0.572} & 0.275 & 0.275 \\
Gemini 3.1 Pro Preview & \underline{0.262} & \underline{0.376} & \textbf{0.510} & \textbf{0.655} & 0.284 & 0.280 \\
Grok 4.3               & 0.251 & 0.364 & 0.413 & 0.560 & \textbf{0.331} & \textbf{0.328} \\
GLM-5V Turbo           & 0.236 & 0.332 & 0.378 & 0.502 & 0.257 & 0.253 \\
\midrule
\textit{Average}       & 0.256 & 0.364 & 0.435 & 0.572 & 0.288 & 0.286 \\
\bottomrule
\end{tabular}
\end{table*}

Table~\ref{tab:headline} reports the five-model headline numbers on the six-lesson intersection (Track 1-1, Track 1-2) and the seven-lesson test split (Track 2).
Three observations stand out.

First, no single model wins all three tracks.
Claude Opus 4.7 leads Track 1-1 (text-only segment coding), Gemini 3.1 Pro leads Track 1-2 (text + frame segment coding), and Grok 4.3 leads Track 2 (lesson-level coverage) on both macro and micro coverage.
This separation between segment-level perception and lesson-level integration is the central empirical finding of \textit{TeachObs} v0.1; a model can dominate one channel of teaching observation without leading on the others.

Second, attaching a single mid-frame image yields a positive macro-F1 vision lift for every model (\S\ref{sec:results-vision-lift}).
The median lift is +0.162 macro and +0.196 micro, and the largest is Gemini 3.1 Pro at +0.247 macro and +0.279 micro.
Even the lowest-performing model in the headline (GLM-5V Turbo) gains +0.142 macro and +0.170 micro from a single frame.
This is consistent with the high visual-code fraction of the coding scheme (20 visual codes out of 39) and, crucially, demonstrates that frontier models can act on that signal directly from a single static image.

Third, GLM-5V Turbo trails on all three tracks, providing an open-frontier baseline that frames the gap between top closed models and a recent open-weights vision LLM.

\subsection{Track 1-1 Text-Only Segment Coding}
\label{sec:results-track-1-1}

Given a 15-second scene transcript in the lesson's source language, the model produces a binary 39-vector over the codes plus a free-text \texttt{reason}.
Claude Opus 4.7 leads Track 1-1 with macro F1 0.277 and micro F1 0.387 on the six-lesson intersection (Table~\ref{tab:headline}).
The best-to-worst gap is 4.1 macro F1 points (Claude 0.277 vs.\ GLM 0.236) and 5.5 micro F1 points (Claude 0.387 vs.\ GLM 0.332).
The four closed frontier models (Claude, GPT-5.5, Gemini, Grok) are tightly clustered within 2.6 macro F1 points of each other, which suggests that on text alone, the frontier-tier ceiling for fixed 15-second classroom segments is around 0.25-0.28 macro F1.

\subsection{Track 1-2 Text + Frame Segment Coding}
\label{sec:results-track-1-2}

Attaching the scene mid-frame as a vision token, while holding the transcript and the prompt identical, reorders the leaderboard.
Gemini 3.1 Pro Preview leads with macro F1 0.510 and micro F1 0.655, ahead of Claude Opus 4.7 (0.438 / 0.570) and GPT-5.5 (0.435 / 0.572), which are now near-tied for second.
Grok 4.3 places fourth (0.413 / 0.560) and GLM-5V Turbo fifth (0.378 / 0.502).
The 1-1 leader (Claude) is overtaken in 1-2 by Gemini, and the macro F1 spread widens from 4.1 to 13.2 points, indicating that the visual channel differentiates frontier models more than the text channel does.

\subsection{Vision Lift, Track 1-2 minus Track 1-1}
\label{sec:results-vision-lift}

Per-model vision lift is computed on the same six-lesson intersection.
Table~\ref{tab:vision-lift} reports the difference $\Delta = $ Track 1-2 $-$ Track 1-1 in macro and micro F1.

Two patterns are worth highlighting.
First, the lift is uniformly positive across all five models, including the open-weights GLM-5V Turbo, even though the visual input is only a single mid-frame image per 15-second scene rather than dense frame sampling.
Second, the lift is largest for Gemini 3.1 Pro at +0.247 macro and +0.279 micro, which is what drives its move into the Track 1-2 leaderboard top.
Vision capability and text-only segment-coding capability are correlated but separable, in the sense that a model can lead under one input condition and not the other.

\subsection{Track 2 Lesson-Level Coverage}
\label{sec:results-track-2}

Track 2 asks the model to produce a lesson-level narrative across eight rated categories (instructional design, instructional delivery, learner response, materials and environment, assessment and closure, lesson strengths, areas for improvement, overall rating), and scores it by reference coverage against the human-derived atomic claim set.
Grok 4.3 leads with macro-mean coverage 0.331 and micro coverage 0.328, ahead of Claude Opus 4.7 (0.295 / 0.293), Gemini 3.1 Pro (0.284 / 0.280), and GPT-5.5 (0.275 / 0.275).
GLM-5V Turbo (0.257 / 0.253) places fifth.

Two implications follow.
First, the Track 2 ranking is not the Track 1 ranking.
Claude is first on Track 1-1 and second on Track 2 cov$_M$, Gemini is first on Track 1-2 and third on Track 2 cov$_M$, and Grok is fourth on both Track 1 modalities but first on Track 2.
Recognizing a teaching action in a 15-second window and writing a lesson-level evaluation that recovers what trained raters wrote are distinct capabilities under the v0.1 evaluation protocol.

Second, absolute coverage is low across the board, with even the leader (Grok 4.3 at 0.331) covering only about a third of the atomic reference claims a trained rater would make.
A floor reading is that lesson-level reference recovery is far from saturated by current frontier models; a ceiling reading is that the single-judge LLM coverage protocol may be conservative, and tighter judge protocols (\S\ref{sec:limitations}) are needed before this gap can be interpreted as a model capability gap.
We return to both readings in the limitations section.

\section{Analysis}
\label{sec:analysis}

The headline numbers in \S\ref{sec:results} show that the three \textit{TeachObs} tracks separate frontier models along different axes, but they do not fully explain those axes.
This section follows a quantitative-qualitative triangulation~\citep{jick1979triangulation} on the same model outputs, with each of three findings reported first as a measurement on the full evaluation split and then illustrated by a representative scene or lesson drawn from manual inspection alongside expert annotations.
Together they give a more specific picture of how current frontier LLMs read classroom video, and where their failure modes are.

\begin{figure*}[!t]
\centering
\includegraphics[width=\linewidth]{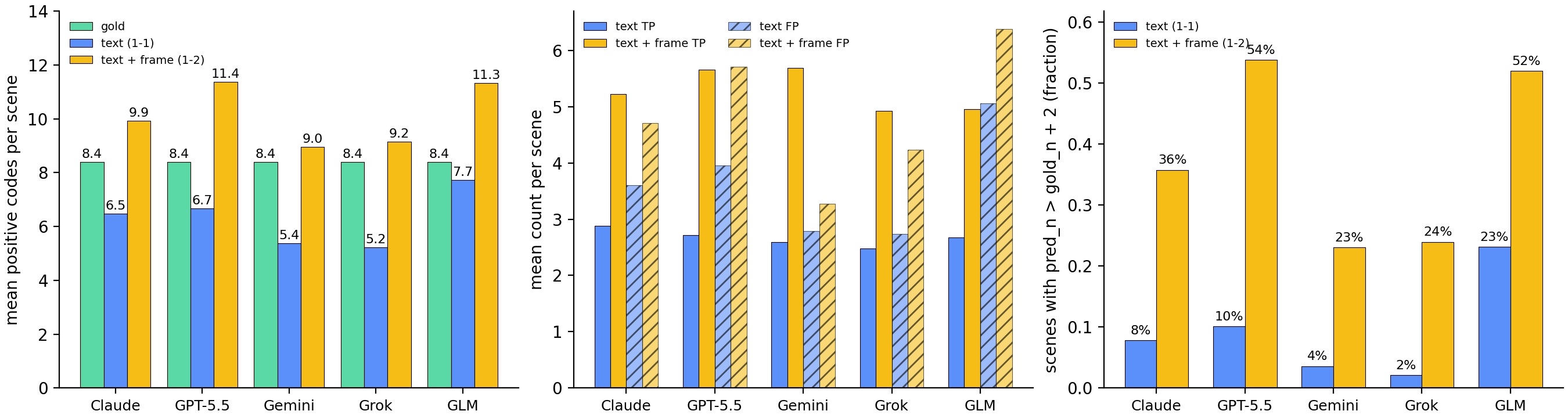}
\caption{Per-model behavior on the six-lesson Track 1 intersection. \textit{Left} reports the mean number of positive codes per scene; gold averages 8.40 codes, text-only predictions under-shoot at 6.30, and text + frame predictions overshoot at 10.15. \textit{Center} decomposes the text and text + frame predictions into TP (solid) and FP (hatched) per scene, showing that the frame lift drags FP up alongside TP. \textit{Right} reports the fraction of scenes with $\text{pred\_n} > \text{gold\_n} + 2$, the rate at which the model overshoots gold by three or more codes.}
\label{fig:analysis-5-1}
\end{figure*}

\begin{figure*}[!t]
\centering
\includegraphics[width=\linewidth]{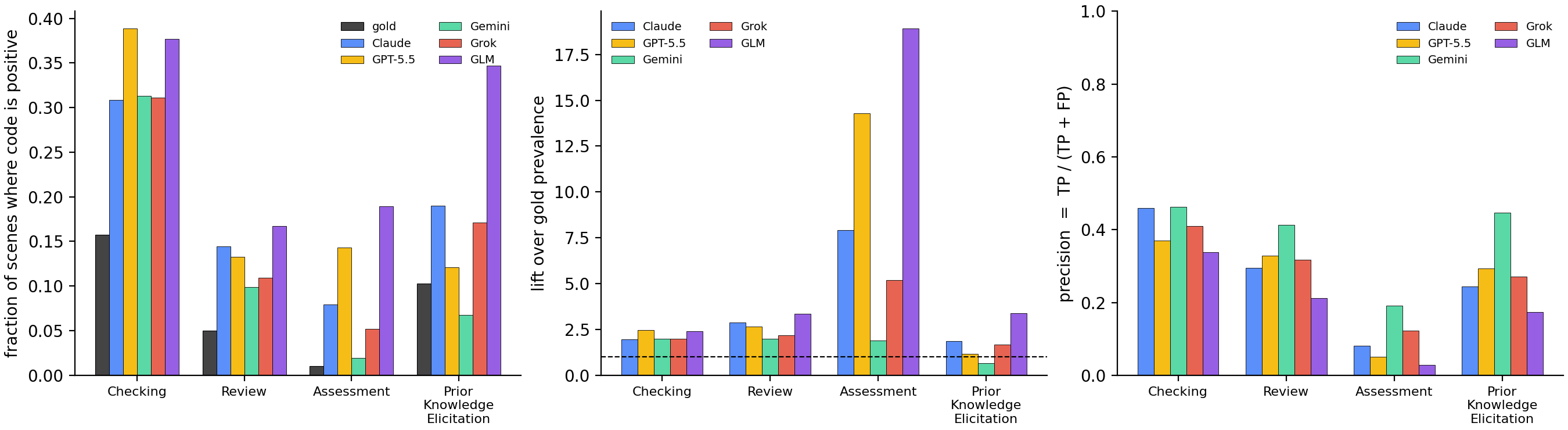}
\caption{Four pedagogical-discourse codes on the six-lesson intersection, text-only Track 1-1. \textit{Left} reports the fraction of scenes where each code is marked positive, with the gold rate next to each model's prediction rate. \textit{Center} reports the over-prediction lift, $\text{pred\_rate} / \text{prevalence}$; the dashed line marks lift one (model matches gold prevalence). \textit{Right} reports precision $\text{TP}/(\text{TP}+\text{FP})$. \textit{Assessment} is predicted at roughly ten times its gold frequency with precision below 0.15 across most models.}
\label{fig:analysis-5-2}
\end{figure*}

\subsection{Text + Frame Strengthens Visual Evidence and Inflates Attribution}
\label{sec:analysis-vision-overattribution}

On the six-lesson Track 1 intersection (1{,}099 scenes), the mean number of positive codes per scene is 8.40 in gold, 6.30 in text-only predictions (under-prediction), and 10.15 in text + frame predictions (over-prediction).
Adding the frame therefore lifts predictions by $+3.85$ codes per scene, of which $+2.62$ land as true positives and $+1.23$ land as false positives, so the vision channel delivers recall with a precision tax.
Figure~\ref{fig:analysis-5-1} reports the per-model decomposition; \textit{left} shows the mean count of positive codes per scene, \textit{center} splits the TP and FP contributions, and \textit{right} reports the fraction of scenes where the model overshoots gold by three or more codes.
The pattern is uniform across all five models, and the FP lift is largest for GLM-5V Turbo and GPT-5.5, the models that already predict the most codes overall.
A per-code pool of $\text{FP}_{\text{text+frame}} - \text{FP}_{\text{text}}$ over the same 1{,}099 scenes localizes the lift; the top contributors are five codes (\textit{Attention Orientation}, \textit{Monitoring}, \textit{Board work}, \textit{Drawing}, \textit{Instructional}), each of which a single mid-frame image plausibly suggests but rarely uniquely licenses.

Manual inspection of the human-expert worksheets makes this pattern concrete.
The case scene S24 scene 191 (47\,m 30\,s to 47\,m 45\,s) has gold with eleven codes; text-only models assign roughly three correct codes, while text + frame models assign seven or more, with several runs exceeding ten codes and introducing labels that do not appear in the human gold.
Adding a single mid-frame image is not free recall.
The skill that separates good text + frame teaching observation from bad is not whether the model can attend to visual signal at all, but whether it can select which of the things it can see are pedagogically load-bearing in the specific 15-second scene at hand.

\subsection{Models Over-Extend Pedagogical Discourse Codes}
\label{sec:analysis-pedagogical-overextension}

For each code we compute its gold prevalence (fraction of scenes marked positive), the per-model prediction rate, and the over-prediction lift $\text{pred\_rate} / \text{prevalence}$.
On text-only inputs, four discourse codes show lifts well above 1.0 with uniformly low precision; \textit{Checking} 2.2$\times$ (precision 0.41), \textit{Review} 2.6$\times$ (0.31), \textit{Assessment} 9.6$\times$ (0.09), and \textit{Prior Knowledge Elicitation} 1.7$\times$ (0.29).
\textit{Assessment} is the extreme case.
In gold only 1.0\% of scenes contain it, but the model average is 9.6\%, and over 90\% of those positives are false positives.
The lift on \textit{Assessment} is uneven across models - GLM 18.9$\times$, GPT-5.5 14.3$\times$, Claude 7.9$\times$, Grok 5.2$\times$, Gemini 1.9$\times$ - and the only model whose \textit{Assessment} predictions are anywhere near gold prevalence is Gemini.
Figure~\ref{fig:analysis-5-2} reports the per-model picture; \textit{left} shows gold prevalence next to each model's prediction rate, \textit{center} reports the over-prediction lift, and \textit{right} reports precision on the same four codes.
The frame channel does not correct this; lifts under Track 1-2 are equal to or larger than under Track 1-1, so the over-extension is a property of the model's pedagogical priors and not its visual perception.

Manual inspection ties the quantitative pattern to a specific scene.
Case scene S2 scene 17 (04\,m to 04\,m 15\,s) has none of the four codes in gold, but several models assign two or more of them under both text and text + frame conditions.
The transcript at that point reads like a generic lesson opener, and the model appears to interpret the entire range of \emph{plausible} script-time pedagogical moves as evidence rather than reading the specific behavior that the observer was tracking.
The four codes share a defining property; each names a teaching move that any well-structured lesson \emph{could} contain, and whose script-time linguistic surface (a question, an instruction, a recap) is generic enough that an LLM reads it as supporting evidence at a much wider range of moments than a trained observer does.

\begin{figure*}[!t]
\centering
\includegraphics[width=\linewidth]{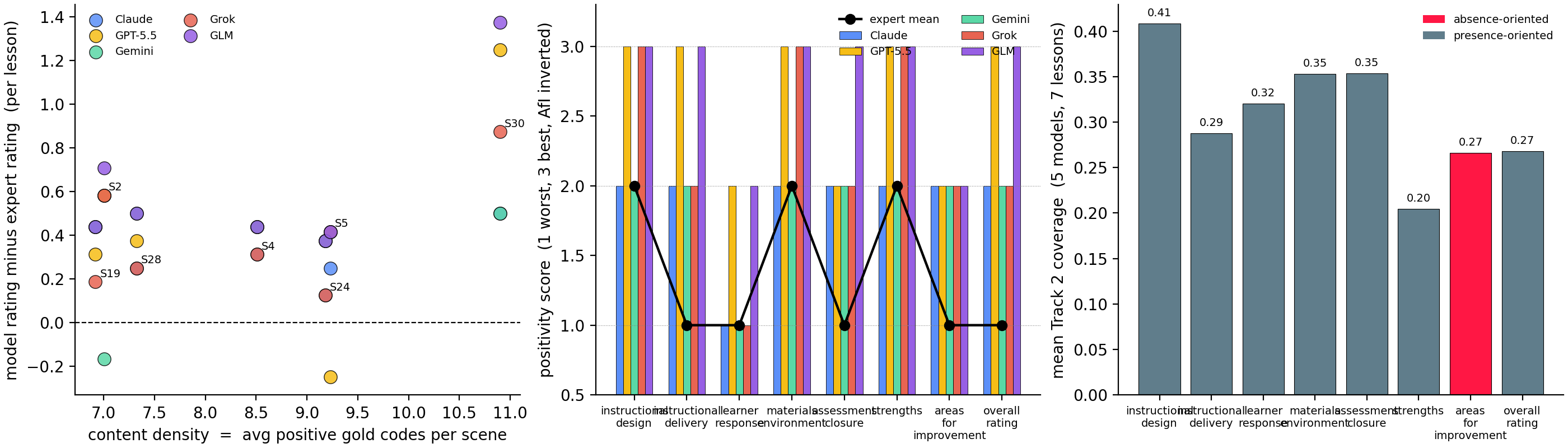}
\caption{Lesson-level over-rating triangulation. \textit{Left} scatters per-lesson over-rating (model minus expert) against content density (mean positive gold codes per scene), one point per (lesson, model); the relationship is positive ($r = +0.53$), and S30 sits at the maximum on both axes. \textit{Center} shows S30 across the eight rated categories; the expert mean (line) sits at or below "Mid" on most categories, while every model places above it. \textit{Right} reports Track 2 mean coverage by category over five models and seven lessons; \textit{Areas for Improvement} is the only absence-oriented category and is consistently lower-covered than the presence-oriented categories aside from \textit{strengths}.}
\label{fig:analysis-5-3}
\end{figure*}

\subsection{Lesson-Level Over-Rating of Procedural and Content-Heavy Lessons}
\label{sec:analysis-lesson-overrating}

We map ratings to a 1-3 positivity score (Low=1, Mid=2, High=3; the \textit{Areas for Improvement} category is inverted because there High means "many improvements needed") and compute, for each lesson, category, and model, the difference $\text{diff} = \text{model} - \text{expert mean}$.
Aggregated across categories and models, the per-lesson over-rating ranges from $+0.25$ (S5) to $+0.90$ (S30).
We then characterize each lesson by two markers computed from the gold Track 1 annotations.
A narrow procedural marker counts positive rates of \textit{Instruction}, \textit{Cueing}, \textit{Organizational Metadiscourse}, and \textit{Attention Orientation}, and the average number of gold codes per scene serves as a content-density proxy.
The narrow procedural marker correlates with per-lesson over-rating at $r = -0.78$, opposite to the qualitative direction, because that narrow set actually tracks interactive checking-style lessons; the content-density proxy correlates positively at $r = +0.53$, and an expanded marker that adds \textit{Content}, \textit{Lecture}, and \textit{Instructional} to the four discourse codes reaches $r = +0.76$.
Figure~\ref{fig:analysis-5-3} integrates these findings; \textit{left} plots content density versus over-rating for every lesson and model, \textit{center} shows the per-category breakdown of expert versus model ratings on S30, and \textit{right} reports Track 2 coverage by category, with \textit{Areas for Improvement} marked separately as the only absence-oriented category.

S30 makes the pattern concrete.
Experts rated S30 Low on instructional delivery, learner response, assessment and closure, and overall, but the five models in our headline set rated it $+0.90$ positivity points above the expert mean on average.
Inspection of model narratives suggests that the lesson's frequent procedural directives ("now do this", "next we'll look at") were read by models as evidence of well-organized instruction, while the trained raters had attended to whether those directives produced learning.
The Track 2 coverage decomposition supports the same reading at the category level; \textit{Areas for Improvement} averages 0.266 across the five models, while the seven presence-oriented categories average 0.314, a small but consistent gap in the same direction.
Lesson-level qualitative evaluation, at least under the v0.1 LLM-as-judge protocol, has a structural bias toward the visible.
A "procedurally clear" lesson reads well to the model because the model's evidence base is what is in the lesson, and current frontier models do not yet pay enough attention to what should have been in the lesson but was not.
This is the second-order pattern the segment-level analysis already hinted at; models are calibrated to attribute pedagogical labels generously when surface evidence is plausible, and that generosity scales from per-scene over-attribution to per-lesson over-rating.

\subsection{Human Validation of Model Outputs on the S2 Sub-Sample}
\label{sec:analysis-human-eval}

On the S2 test lesson, three expert raters (H1-H3) checked Track 1 model outputs on the first 30 scenes under both Track 1-1 (text) and Track 1-2 (text + frame) using a five-stage scene-level verdict, and three further raters (H4-H6) reviewed the single LLM judge's covered or missed call on 6 S2 atomic reference claims $\times$ 5 models; the sample, rubric, and statistical tests in Table~\ref{tab:human-eval-summary} were pre-registered before the returned xlsx files were inspected.
At scene level, adding the mid-frame reduces the human-judged wrong rate from $33.3\%$ to $17.8\%$ (McNemar exact $p = 0.013$ over 90 paired scenes), which aligns the human verdict with the macro F1 lift in \S\ref{sec:analysis-vision-overattribution} and constrains the over-attribution reading there - the true-positive gain dominates the false-positive growth in the holistic scene judgment on this sample, even though the per-code FP arithmetic still rises; inter-rater agreement on the five-stage scale is modest under text (Fleiss $\kappa = 0.20$) and chance-level under text + frame ($\kappa = -0.11$), which we read as a property of the holistic verdict task rather than of the model outputs.
At claim level, the human majority agrees with the single LLM judge on $30/30$ (model, claim) pairs (Wilson lower bound $88.6\%$), with no directional bias detectable in the few rater-cell disagreements (binomial $p = 1.0$), so the single-judge protocol used in \S\ref{sec:results-track-2} and \S\ref{sec:analysis-lesson-overrating} is corroborated by independent expert raters at the limit of what 30 pairs can resolve; a full-test-split version of this analysis and a family-disjoint three-judge majority for Track 2 are scheduled for the next release.

\begin{table}[t]
\centering
\footnotesize
\setlength{\tabcolsep}{4pt}
\caption{Human re-judgment statistics on the S2 sub-sample. Track 1 is computed over 30 scenes $\times$ 3 raters $\times$ 2 modalities (180 ratings, raters H1-H3). Track 2 is computed over 30 (model, claim) pairs $\times$ 3 raters (90 cells, raters H4-H6).}
\label{tab:human-eval-summary}
\begin{tabular}{@{}lr@{}}
\toprule
\textbf{Quantity} & \textbf{Value} \\
\midrule
\multicolumn{2}{@{}l}{\textit{Track 1-1 (text), 30 scenes $\times$ 3 raters}} \\
\quad Wrong rate (Wilson 95\% CI)            & 33.3\% [24.5, 43.6] \\
\quad Fleiss $\kappa$ 5-class (95\% CI)      & 0.20 [-0.02, 0.42] \\
\quad Cohen $\kappa$ pairwise (range)        & -0.03 to 0.57 \\
\midrule
\multicolumn{2}{@{}l}{\textit{Track 1-2 (text + frame), 30 scenes $\times$ 3 raters}} \\
\quad Wrong rate (Wilson 95\% CI)            & 17.8\% [11.2, 26.9] \\
\quad Fleiss $\kappa$ 5-class (95\% CI)      & -0.11 [-0.24, 0.03] \\
\quad Cohen $\kappa$ pairwise (range)        & -0.05 to 0.04 \\
\midrule
\multicolumn{2}{@{}l}{\textit{Track 1 paired (text vs vision wrong, 90 paired scenes)}} \\
\quad McNemar exact $p$                      & 0.013 ($b{=}21$, $c{=}7$) \\
\midrule
\multicolumn{2}{@{}l}{\textit{Track 2 (30 pairs $\times$ 3 raters, S2 claims)}} \\
\quad Human-majority vs judge agreement      & 30/30 = 100\% [88.6, 100] \\
\quad Rater-cell agreement (95\% CI)         & 85/90 = 94.4\% [87.6, 97.6] \\
\quad Disagreement direction (cov./missed) $p$ & 1.0 (2 / 3) \\
\bottomrule
\end{tabular}
\end{table}

\section{Conclusion}
\label{sec:conclusion}

We presented \textit{TeachObs}, a multimodal teaching observation benchmark built on 30 public K-12 lesson videos from eight countries.
The dataset pairs two reference layers on the same lessons - a segment-level multi-label gold over 5{,}158 fifteen-second scenes annotated for 39 binary observation codes, and lesson-level qualitative narratives across ten evaluation categories produced by three expert raters - and ships with the operational definitions, reliability statistics, and prevalence-aware aggregation rules used to construct the gold.
On the v0.1 test split (seven lessons, 1{,}030 atomic reference claims), five vision-capable frontier LLMs separate along distinct axes; Claude Opus 4.7 leads text-only segment coding, Gemini 3.1 Pro leads segment coding with the mid-frame attached, and Grok 4.3 leads lesson-level coverage.
Three findings from the analysis section indicate that the dataset measures more than headline F1.
Adding a single mid-frame image inflates per-scene attributions alongside genuine recall; models over-extend pedagogical-discourse codes such as \textit{Assessment} by an order of magnitude beyond their gold prevalence; and at the lesson level, models over-rate procedurally clear or content-heavy lessons and under-cover what trained raters write about what was \emph{not} happening.
\textit{TeachObs} is intended as a research resource on classroom teaching, and is not designed as an instrument for evaluating individual teachers.
The release at acceptance time will be tied to a persistent identifier together with the prompts, scoring scripts, and human qualitative-analysis worksheets used to produce the numbers in this paper.

\section*{Limitations and Ethical Considerations}
\label{sec:limitations}
\label{sec:ethics}

\textit{TeachObs} spans eight countries but a substantial share of the 30 lessons are Korean, and all human annotations were produced by Korean educational researchers, so the labels do not automatically generalize beyond this training context.
Lesson-level rater coverage is asymmetric (R1, R2 = 30 lessons; R3 = 10), Track 1-2 mid-frame attachments are missing for S4 so Track 1 is reported on the six-lesson intersection, and Track 2 coverage uses a single LLM judge (\texttt{openai/gpt-5-mini}) - a family-disjoint three-judge majority is scheduled.
The human re-judgment in \S\ref{sec:analysis-human-eval} is on a pre-registered S2 subsample; the within-S2 statistics are powered for the effects reported but lesson-level generalization awaits the next release.

The source recordings are publicly available; we redistribute metadata, transcripts, and segment specifications but not the video frames, and downstream users should apply additional masking (e.g., face blurring) when redistributing derived artifacts.
\textit{TeachObs} is a research resource on classroom teaching and is explicitly not designed or appropriate as an instrument for evaluating individual teachers.
Annotators and expert raters were trained classroom-observation researchers who gave informed consent, could withdraw without penalty, were compensated at the host institution's standard rate, and did not handle student records or private interaction logs.
Icons in Figure~\ref{fig:overview} are by Creative Stall Premium, Freepik, Hilmy Abiyyu A., and Uniconlabs from \url{www.flaticon.com}.

\section*{Use of Generative AI}
\label{sec:genai}

We used Anthropic's Claude Opus models (versions 4.5, 4.6, 4.7) for writing polish and code edits under direct human review; the research ideas, benchmark design, coding scheme, annotation procedure, analysis, and conclusions were produced by the human authors, and every passage, table, figure caption, and number reported here was verified by a human author before submission.
Parts of Figure~\ref{fig:overview} were produced with ChatGPT Image 2.0.

The five frontier LLMs evaluated on \textit{TeachObs} are themselves generative AI systems (listed with their OpenRouter slugs in \S\ref{sec:results-setup}).
Two further generative AI components participate in the evaluation pipeline rather than in the writing - the Track 2 atomic-claim decomposer uses GPT-5.5 (\texttt{openai/gpt-5.5}) and the Track 2 LLM-as-judge uses GPT-5-mini (\texttt{openai/gpt-5-mini}); Whisper large-v3 was used in an earlier stage of transcript construction and the v0.1 release replaces those automatic transcripts with the cleaned TSV transcripts from the original coder workflow.



\clearpage
\appendix

\end{document}